\documentclass[final]{cvpr}

\usepackage{array}
\usepackage{booktabs}
\usepackage{times}
\usepackage{epsfig}
\usepackage{graphicx}
\usepackage{amsmath}
\usepackage{amssymb}
\usepackage{bm}

\usepackage[pagebackref=true,breaklinks=true,colorlinks,bookmarks=false]{hyperref}

\def\cvprPaperID{****} 
\def\confYear{CVPR 2021}

\begin{document}


\title{Single-Training Collaborative Object Detectors\\Adaptive to Bandwidth and Computation}

\author{Juliano S. Assine\textsuperscript{1} ~~~J. C. S. Santos Filho\textsuperscript{2} ~~~Eduardo Valle\textsuperscript{1}\\ \\
\textsuperscript{1}RECOD Lab. ~~~~ \textsuperscript{2}WissTek Lab.\\School of Electrical and Computing Engineering (FEEC) \\ University of Campinas (UNICAMP), Brazil\\

{\tt\small jsiloto@dca.fee.unicamp.br, candido@decom.fee.unicamp.br, evalle@unicamp.br}
}

\maketitle





\maketitle

\begin{abstract}
   In the past few years, mobile deep-learning deployment progressed by leaps and bounds, but solutions still struggle to accommodate its severe and fluctuating operational restrictions, which include bandwidth, latency, computation, and energy. In this work, we help to bridge that gap, introducing the first configurable solution for object detection that manages the triple communication-computation-accuracy trade-off with a single set of weights. Our solution shows state-of-the-art results on COCO–2017, adding only a minor penalty on the base EfficientDet-D2 architecture. Our design is robust to the choice of base architecture and compressor and should adapt well for future architectures.
\end{abstract}


\section{Introduction}
\label{sec:introductio}
Mobile and embedded devices comprise most computing devices, both in raw number and market share~\cite{embedded_market_by_2019}. The impact of bringing to them cutting-edge computer vision models can hardly be overstated, given their ubiquity.

Deep neural networks not only solidified core tasks of computer vision — such as image classification and object detection — but enabled many applications not even considered until then — such as style transfer, pose transfer, and natural image generation, to name a few. Much of current research focuses on the real-world deployment of those solutions. Even smaller models, like image classifiers, pose considerable challenges for deployment in mobile and embedded devices, while larger models, like object detectors, have comparatively lagged. 


A crucial trend in improving that scenario is \textit{collaborative intelligence}~\cite{bajic2021collaborative}, which addresses dividing deep learning inference across multiple devices. In particular \textit{split computing}~\cite{matsubara2020split}  computes the first layers of a deep model in the local, mobile device while delegating the last layers to remote servers.

\begin{figure}
\begin{center}
   \includegraphics[width=0.99\linewidth]{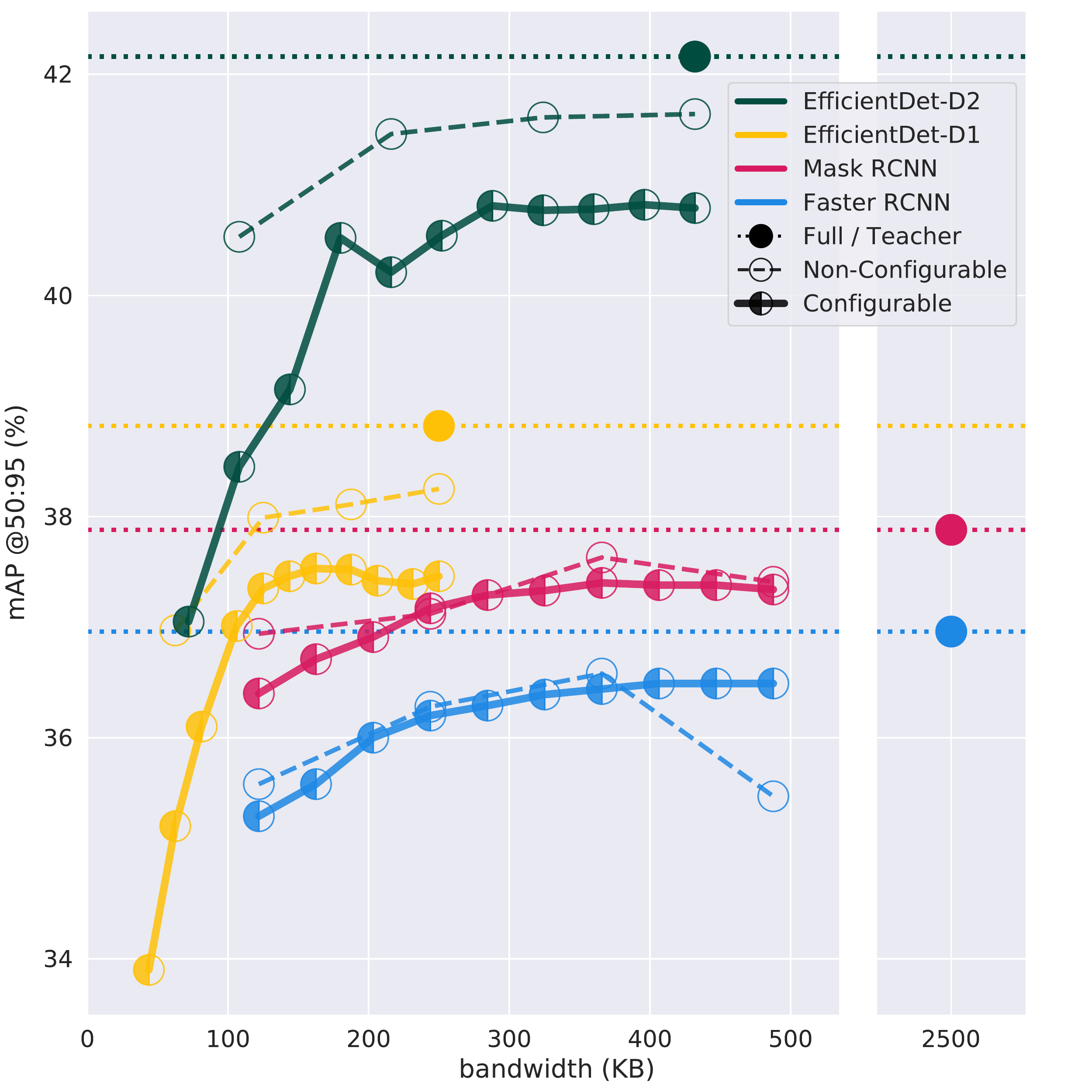}
\end{center}
   \caption{Visual summary of contributions. Networks with full features (full dots, horizontal dotted lines) provide baseline performance for object detection. Previous works~\cite{matsubara2021neural} distilled those networks into encoders with smaller features (magenta and blue, hollow dots, dashed lines). Our work explores both enhanced architectures (dark green, gold), and \textit{configurable} encoders, allowing on-the-fly adaptation with a single set of weights (halved dots, solid lines). 
   The task is object detection on COCO–2017.}
\label{fig:graph_bw_map}
\end{figure}


While promising, those ideas have yet to surmount the severe technical constraints of mobile devices, which include limitations of network bandwidth and latency, connection reliability, computational power, energy consumption, thermal dissipation, etc. Before becoming practical for widespread consumer deployment, solutions will have to become adaptable to multiple, realistic constraints. 

Many “adaptive” deep learning models require separate training and weight-sets for each expected runtime condition. That imposes a severe compromise on the number of settings to which the model may adapt. Worse: separate weight-sets limit the responsiveness of the system and increase power consumption. 




The \textbf{main contribution} of this work is breaking that compromise by proposing a \textit{configurable split object detector} with a single set of weights. Our model is the first full-neural (i.e., without non-differentiable layers) split object detector adaptable for bandwidth with a single set of weights. It is the first split computer vision model flexible on-the-fly for both bandwidth and computation. Another contribution is an analytical survey of the recent literature on split models for computer vision.

The text follows the usual organization, with literature survey next, proposed technique in Section~\ref{sec:design}, experiments in Section~\ref{sec:experiments}, and conclusions/discussion last.




\section{Related work}
\label{sec:related_work}
\begin{table*}[]
\begin{tabular}{lc>{\raggedright}p{7.8cm}>{\raggedright}p{3.7cm}c}
\toprule
Ref$_\mathrm{Year}$ & Tasks & Techniques / Architectures & Datasets & Conf$_\mathrm{BW}$ \\
\midrule
\cite{luo2018deepsic}$_{2018}$ & C AE & Training from scratch / Custom Architecture & ImageNet & No \\
\cite{eshratifar2019towards}$_{2019}$ & C & Training from scratch / ResNet-50 & ImageNet (subset) & No \\
\cite{eshratifar2019bottlenet}$_{2019}$ & C & Compression-augmented training (JPEG) / ResNet & ImageNet (subset) & Yes \\
\cite{assine2019compressing}$_{2019}$ & C & PCA /  MobileNet-v2 (0.25-1.0) & ImageNet & Yes \\
\cite{matsubara2019distilled}$_{2019}$ & C & Head network distillation / DenseNet (169 \& 201), ResNet-152, Inception-v3 & Caltech-101 & No \\
\cite{shao2020bottlenet++}$_{2020}$ & C & Fine-tuning / ResNet & CIFAR-100 & No \\
\cite{matsubara2020head}$_{2020}$ & C & Fine-tuning, Knowledge distillation, Head network distillation / MobileNet-v2 (1.0), MnasNet (0.5, 1.0), DenseNets (169, 201), ResNet-152, Inception-v3 & ImageNet & No \\
\cite{shao2020communication}$_{2020}$ & C & Fine-tuning, Structured pruning / ResNet-18 & CIFAR-10 & No \\
\cite{hu2020fast}$_{2020}$ & C & Fine-Tuning / MobileNet-v2 (1.0) & CIFAR-10, CIFAR-100 & No \\
\cite{jankowski2020joint}$_{2020}$ & C & Fine-tuning, Structured pruning / VGG-16  & CIFAR-100 & No \\
\midrule
\cite{choi2018deep}$_{2018}$ & OD & Compression-augmented training (HEVC) / YOLO-9000-v2 & VOC$_{2007}$ & Yes \\
\cite{choi2018near}$_{2018}$ & C OD AE & Custom deep feature CODEC / YOLO-v2, Darknet-19$_{448}$, VGG-16, ResNet & VOC$_{2007}$ & No \\
\cite{chen2019lossy}$_{2019}$ & C OD IR & HEVC Compression without retraining / VGG-16, ResNet (50, 101, 152), Faster~RCNN & ImageNet (subset), PKU VehicleID, VOC$_{2007}$ & Yes \\
\cite{choi2020back}$_{2020}$ & OD & Channel selection + Reconstructive decoder / YOLO-v3 & COCO$_{2014}$  & No \\ 
\cite{cohen2020lightweight}$_{2020}$ & OD & Entropy-constrained quantizer / YOLO-v3 & COCO$_{2017}$ \textit{mAP@50} & Yes \\
\cite{matsubara2020split}$_{2020}$ & OD IS & Head network distillation / Faster~RCNN, Mask~RCNN & COCO$_{2017}$ \textit{mAP@50:95} & No \\
\cite{matsubara2021neural}$_{2020}$ & OD IS KD & Generalized head network distillation / Faster~RCNN, Mask~RCNN & COCO$_{2017}$ \textit{mAP@50:95} & No \\ 
\bottomrule
\end{tabular}
\caption{Summary of the state of the art on split computing featuring lossy compression. Tasks: Auto Encoder, Classification, Image Retrieval, Instance Segmentation, Keypoint Detection, Object Detection. Conf$_\mathrm{BW}$ indicates works configurable in the sense of Section~\ref{sec:configurability}. There is a surge of interest in the subject, but the diversity of tasks, datasets, and metrics makes comparison across works challenging.}
\label{table:sota}
\end{table*}
In this section, we focus on the recent literature on split computing without aiming at an exhaustive review of collaborative intelligence.  Bajić \etal~\cite{bajic2021collaborative} compiled a very recent summary of the challenges in collaborative intelligence, while Wang \etal~\cite[Section~V]{wang2020convergence} and Shi \etal~\cite{shi2020communication} organized comprehensive surveys of deep learning in edge computing, including collaborative intelligence. Table~\ref{table:sota} summarizes the works within our scope: those that, at inference time, split the deep-learning network, adding lossy feature compression/decompression between the two halves. Works below the mid-line, which address object detection, are particularly related to ours. 



Early works tended to focus on the feasibility of the concept, showing that split computing outperformed compressing the input image and offloading the inference entirely to a remote server~\cite{assine2019compressing, choi2018deep, choi2018near, eshratifar2019towards, luo2018deepsic}. In terms of bandwidth, the split solution was a strong contender, since unsupervised deep compression was already surpassing classical compression~\cite{balle2016end}, but its on-device computational feasibility was far from clear. Eshratifar \etal  ~\cite{eshratifar2019bottlenet} drew attention with a simple framework of autoencoding+JPEG compression analyzed under a power consumption model for wireless networks~\cite{huang2012close}, showing clear gains over compressed input offloading in terms of bandwidth, latency, \textit{and} device power consumption. 

In contrast, recent split architectures for object detection show little concern for measuring computational costs, focusing on compressing the features and reducing bandwidth. That choice reflects the challenge of making split computing feasible for the task, as networks for object detection tend to branch on upper layers, creating a huge amount of communication. All existing solutions cope with that phenomenon by splitting the network before the branching stages (Figure~\ref{fig:split_object_detectors}). While Matsubara \etal~\cite{matsubara2021neural} proposes a framework for distilling the encoders of split networks, other works focus on lightweight compression that forgoes retraining~\cite{ choi2020back, cohen2020lightweight}. 

As shown Table~\ref{table:sota}, even when existing works share the same task, the lack of standardization hinders direct comparison. In addition, many works assume unrealistic settings for the scenario of actual mobile applications (see~\cite[Section~3.2]{matsubara2020split} for a lucid critique in that direction). Aside from Matsubara \etal~\cite{matsubara2021neural, matsubara2020split}, Cohen \etal~\cite{cohen2020lightweight} are the only other authors known to use COCO-2017, but even among that limited set of works, lack of agreement on the metrics prevents direct comparison. In this work, we follow Matsubara \etal, choosing the mAP@50:95 as a more challenging (and more robust) metric for object detection instead of the easier (and potentially noisier) mAP@50. Please refer to Section~\ref{sec:materials_and_methods} for an explanation of the metric.

\subsection{Configurability}
\label{sec:configurability}

An issue with existing art is the lack of standard terminology. In this work, we define \textbf{configurable} models as those able to change operational parameters (e.g., bandwidth) without needing a separate set of weights for each setting — at least on the \textbf{encoder} (the “half” of the model that resides on the local device). Let the reader beware that literature employs the terms “adaptive”, “dynamic”, “flexible”, and others with an array of meanings, ranging from relatively rigid models, trained separately (with a different weight-set) for each operational condition, until highly flexible models, able to select different computation paths on-the-fly based on each input (e.g., using early exits)~\cite{teerapittayanon2016branchynet, yang2020resolution}.

Multiple sets of weights, in addition to storage costs, hinder the model’s time performance, as models must stop for potentially several inference cycles to load the weights to memory~\cite{luo2020comparison}. Weight reloading may also severely impact battery life, as memory transfer and storage access use orders of magnitude more power than arithmetic computation (see~\cite{horowitz20141} \textit{apud}~\cite{yang2017designing}).

The simplest way to provide \textbf{bandwidth configurability}, i.e., the ability to adapt the size of the representation between the encoder and the \textbf{decoder} (the “half” of the model that resides on the remote server) is to have an adjustable compressor. Straightforward approaches such as Principal Component Analysis~\cite{assine2019compressing} are easily parameterizable on-the-fly, but tend to discard important information (e.g., the spatial structure of tensors). In contrast, optimized quantization schemes~\cite{cohen2020lightweight} are very computation-efficient but have a limited ability of compressing representations on their own, constituting only one piece among others for most codec pipelines.


On the opposite end of the sophistication spectrum, full-fledged image and video compression schemes, like JPEG and HVEC, may be employed on the features. The compressor may be introduced only during inference time, e.g., by applying HEVC to the output of the convolutional layers, treating the collection of feature maps output by the channels of a layer across a given batch as if they were successive temporal frames~\cite{chen2019lossy}. Instead, the training may be compression-augmented / compression-aware, e.g., by introducing JPEG~\cite{eshratifar2019bottlenet} or HVEC~\cite{choi2018deep} as a non-trainable layer, and using several quality configurations as a form of regularization. 

Instead of introducing an extraneous compressor, many split architectures choose a learned neural-network compressor/decompressor (i.e., an auto-encoder) that can be trained seamlessly with the rest of the network~\cite{choi2018deep, eshratifar2019bottlenet, eshratifar2019towards,hu2020fast, jankowski2020joint,matsubara2019distilled, matsubara2020head, matsubara2021neural, matsubara2020split, shao2020bottlenet++, shao2020communication}. In addition to such conceptual attractiveness, learned compressors have been shown to outperform, in terms of bandwidth savings, classical compression~\cite{balle2016end}. The challenge is making those full-neural schemes configurable. Existing schemes (e.g., ~\cite{hu2020fast}) often focus on optimizing data transmission for different bandwidth regimens, but require different sets of weights. One exception is the work of Choi \etal~\cite{choi2020back}, which allows selecting a subset of channels on the same encoder, but needs different decoders (with different weights) for the reconstruction. At the time of this writing, we are not aware of any full-neural architecture configurable on both the encoder and the decoder. 



In comparison to bandwidth, \textbf{computational configurability} has not yet been explored by the literature, even though non-configurable options have tackled the impact of computation for split classification models~\cite{eshratifar2019bottlenet, jankowski2020joint, shao2020bottlenet++, shao2020communication}. Theoretical results on information bottlenecks~\cite{tishby2015deep, achille2018emergence}, suggesting that the \textit{actual} information content of deep representations decreases the further we advance layer-wise, motivated the idea of a computation–communication tradeoff: the client may save communication by advancing to further layers and obtaining more compressible representations. That compromise may be exploited to obtain optimal split-points, but so far, the proposed solutions require different trained weights for each choice of split-point.

In a more applied direction, the seminal work of slimmable networks~\cite{Yu2019}, in which the network is trained simultaneously with different subsets of the channels, as well as follow-up models~\cite{cai2020once, yang2020mutualnet, yu2019universally}, provided a mechanism to achieve configurability on deep networks without requiring multiple split points. As far as we know, our work is the first to employ that family of techniques on split networks.





\subsection{Other directions}

We mention, for the sake of curiosity, interesting directions taken by recent works, which include multiple split-points and multiple representations at the split point~\cite{alvar2020bit,eshratifar2019jointdnn}, allocation policies for split computing~\cite{callegaro2020optimal}, resilience to network packet loss ~\cite{liu2020improve}, and inference-time privacy~\cite{bajic2021collaborative}. We mention, in particular, collaborative intelligence under additive white Gaussian noise channels~\cite{jankowski2020joint, shao2020bottlenet++,shao2020communication}, which optimize learning, compression, and coding all at once, and strive to minimize on-device computation and create a standardized comparison, even if current works are limited to the setting of 32$\times$32 images of CIFAR.


\section{Proposed split design}
\label{sec:design}
\begin{figure*}[htb]
\begin{center}
\includegraphics[width=0.95\linewidth]{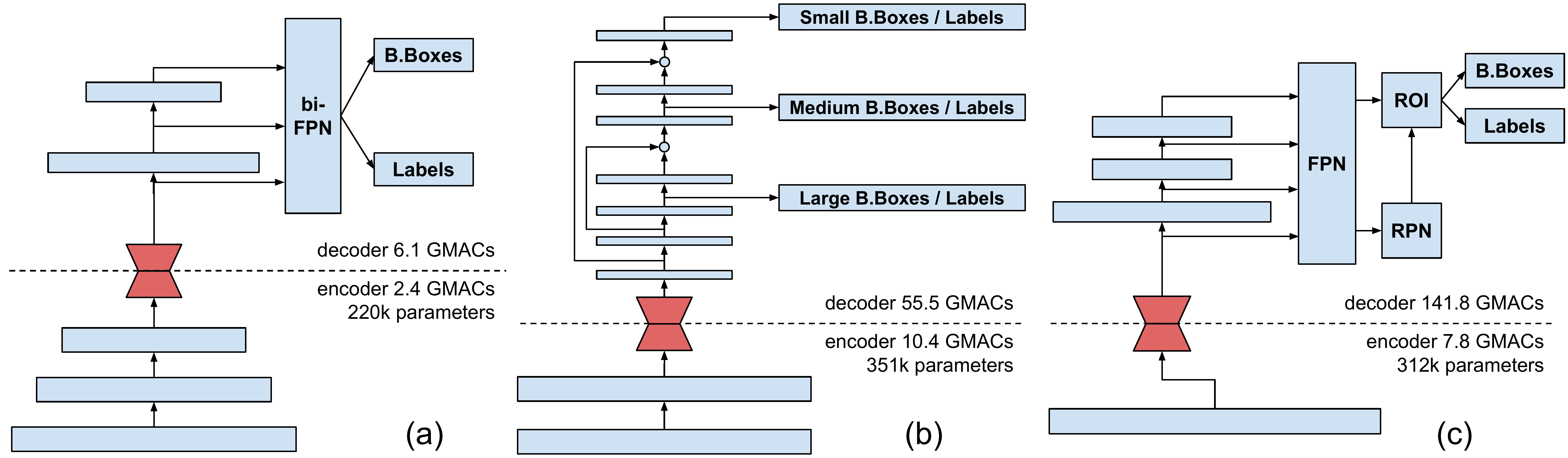}
\end{center}
   \caption{Three split architectures for object detection: (a) one of ours, based on EfficientDet–D2, (b) one by Cohen et al.~\cite{cohen2020lightweight} based on YOLO–v3—Darknet, (c) one by Matsubara et al.~\cite{matsubara2020split} based on Mask~RCNN—ResNet. The encoder is to run on the mobile client, and the decoder, on a remote server. Those models require a compressor/decompressor (in red), and their \textit{bandwidth} is the size of the feature that leaves the compressor. The mAP@50:95 of the full (teacher) networks on COCO-2017 validation set was (a) 42.2, (b) 37.9 and (c) 38.0.  }
\label{fig:split_object_detectors}
\end{figure*}

\begin{figure}
\begin{center}
   \includegraphics[width=0.99\linewidth]{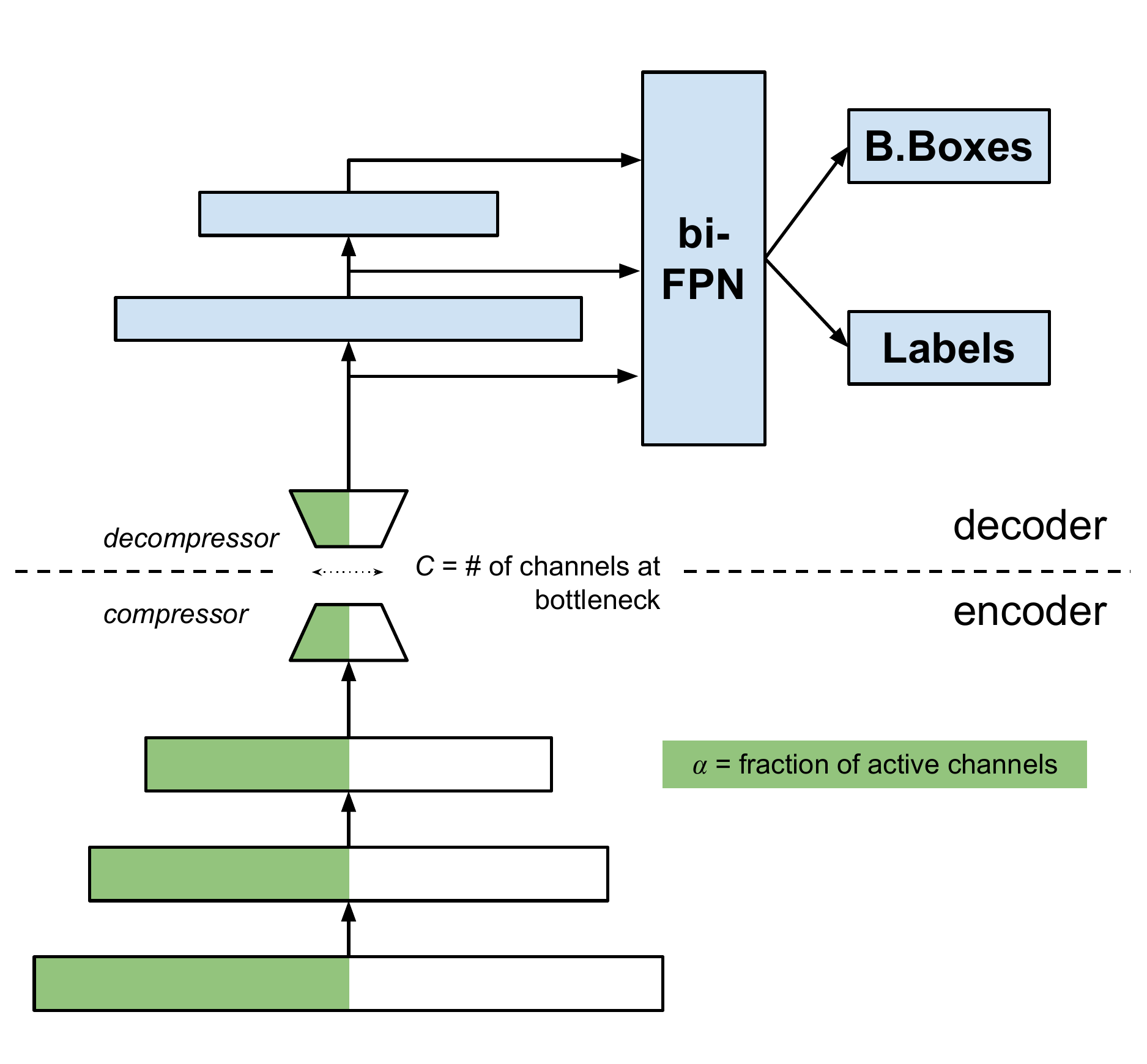}
\end{center}
   \caption{Our architecture follows the scheme shown in Figure~\ref{fig:split_object_detectors}, but introduces configurability of both bandwidth and computation with $\alpha$, and a (fixed) choice $C$ of maximum bandwidth.} 
\label{fig:proposed_architecture}
\end{figure}

As seen in the previous section, split deep-learning models appear as a promising solution for implementing costly inference on mobile devices, but introduce a communication bottleneck between the local encoder and the remote decoder that must adapt to a varying bandwidth availability. For object detection, the split happens before the branching of the network, and on full-neural solutions, we introduce a network block at the end of the encoder to compress the representation and a corresponding reconstruction block in front of the decoder (Figure~\ref{fig:split_object_detectors}).

The main novelty of the proposed architecture (Figure~\ref{fig:proposed_architecture}) is its \textit{reconfigurability}, in the sense explained in Section~\ref{sec:configurability}. All models we introduced are \textit{bandwidth-reconfigurable}, and several are \textit{computation-reconfigurable} (in the encoder) as well. Another novelty was the use of the high-performance EfficientDet~\cite{tan2020efficientdet} as foundation for the architeture, which entailed the need for a compressor compatible with those innovations. We discuss each of those innovations below.


\paragraph{Configurability.} Existing works achieve bandwidth flexibility either by introducing non-differentiable operations into the neural network or by reloading a set of network weights trained for each transmitted representation size. The advantage of a full-neural solution is that the compression adapts to the data, since it is integrated seamlessly into the model. However, existing full-neural models require a different set of weights for each bandwidth configuration, creating a compromise between flexibility (configuration points) and training/space cost (number of weight-sets). Even worse, those models introduce a disrupting latency during reconfiguration, each time a weight-set must be loaded to memory.

In contrast, our solution makes the model adaptive with a single set of weights. Bandwidth reconfiguration becomes extremely flexible because increasing the number of points does not require adding more weight-sets. Changing operational constraints does not require reloading weights: it is achieved by dynamically selecting $\pmb\alpha$, \textbf{the fraction of active channels} in the convolutional layers (Fig.~\ref{fig:proposed_architecture}).

That flexibility is gained thanks to slimmable networks~\cite{Yu2019}, a set of training methods in which a single set of weights learns to perform well for multiple channel-widths (values of $\alpha$). As base (teacher) architectures, we test Mask RCNN and the Faster RCNN, for the sake of comparison with Matsubara \etal~\cite{matsubara2021neural, matsubara2020split}, and introduce EfficientDet. For each architecture, we must choose the slimmable layers (discussed below) and the layers to use on the distillation loss: we used the outputs of blocks 1–4 on Faster RCNN and Mask RCNN, and blocks 3–5 on EfficientDet, chosen because they are the branching points of the network. We present the training procedure in detail in Section~\ref{sec:training}. 






\begin{figure*}[htb]
\begin{center}
\includegraphics[width=0.99\linewidth]{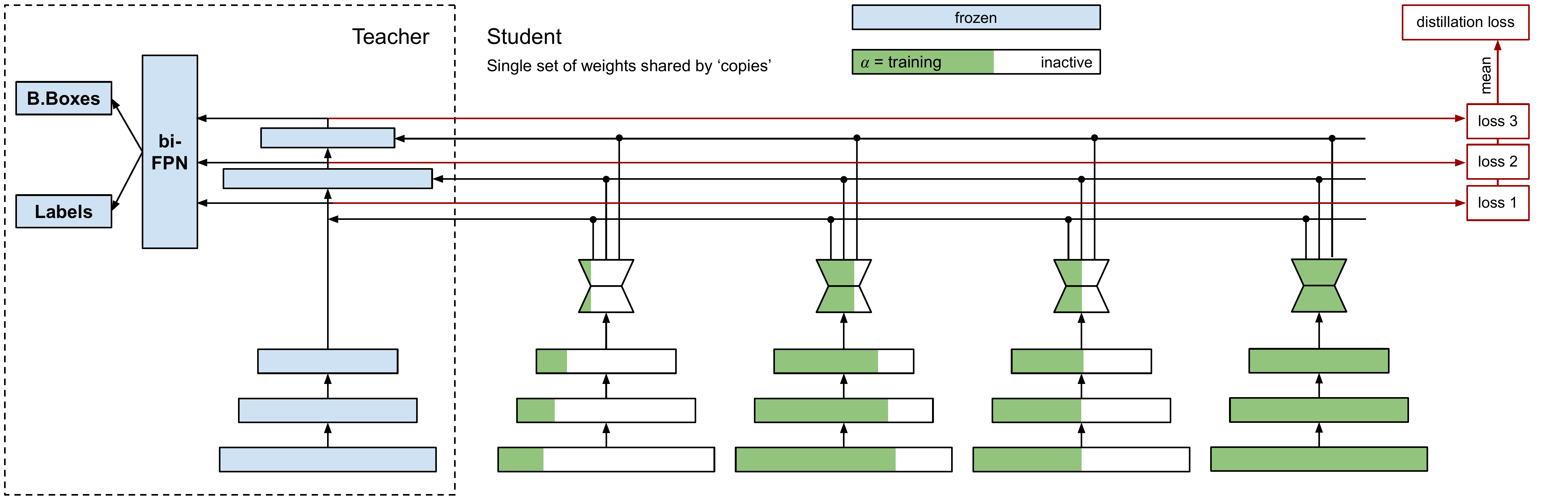}
\end{center}
   \caption{Training of the proposed split slimmable architecture. First the reference teacher architecture (non-split, non-slimmable) is trained and frozen. Then the slimmable architecture is learned by distillation: $N$ “copies” of the model (all sharing a single set of weights) are trained concurrently, for varying number of active channels.}
\label{fig:training_method}
\end{figure*}

\begin{figure}
\begin{center}
  \includegraphics[width=0.95\linewidth]{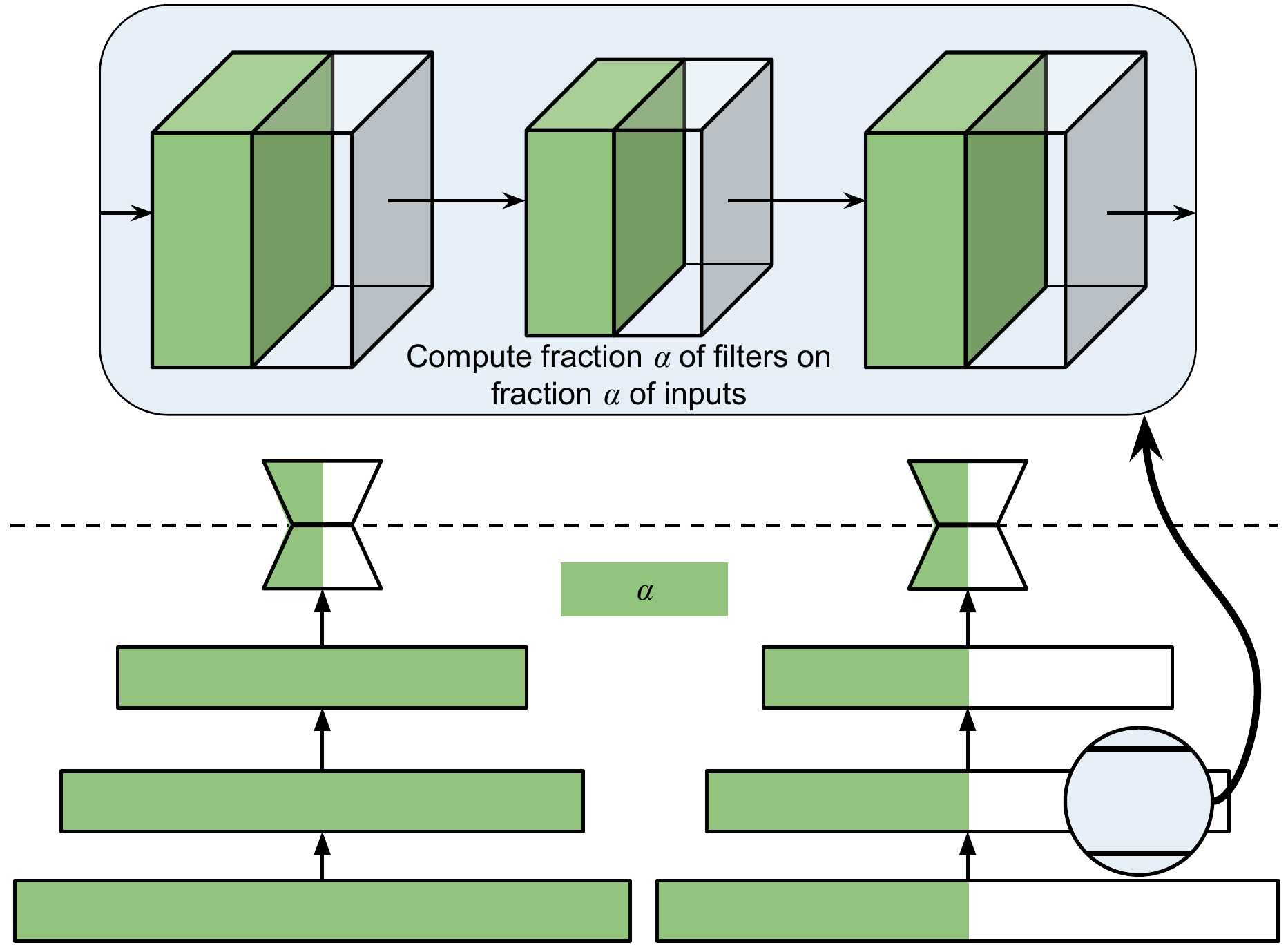}
\end{center}
    \caption{Encoder design: depending on the choice of the slimmable layers, we may achieve bandwidth-only configurability (left), or both bandwidth and computation configurability (right). In the latter, at each internal convolutional layer, $\alpha$ has a quadratic effect on computation, by reducing both input and output.}
\label{fig:slimmable_bottleneck}
\end{figure}

\paragraph{Computation-configurability.} Depending on the choice of the slimmable layers, we may achieve both bandwidth and computation configurability, or only the former (Figure~\ref{fig:slimmable_bottleneck}). 

For bandwidth-only configurability, we only have to make the compressor and decompressor slimmable (technically, there will be some impact in computation, but too small to be of practical use). For configurability in both criteria, we may elect to make all convolutions in the encoder slimmable. Because both computation and communication are critical resources in mobile devices, those designs may be unavoidable to regulate energy and thermal constraints.

While the impact of varying a single convolutional layer is small, when $\alpha$ (explained above) is applied to neighboring layers, it accumulates \textit{quadratically}, affecting both the output of one layer, and the input of the next:

\vspace{-0.5cm}
\begin{equation}
\mathit{MAC} = k^2\cdot \alpha c_\mathrm{in} \cdot \alpha c_\mathrm{out} = k^2\cdot \alpha^2 \cdot c_\mathrm{in} \cdot c_\mathrm{out}
\vspace{-0.1cm}
\end{equation}

\noindent where, for a given layer, $\mathit{MAC}$ is the number of \textit{m}ultiply-\textit{ac}cumulate arithmetic operations, $k$ is the kernel width, $c_\mathrm{in}$ is the number of input channels, and $c_\mathrm{out}$ is the number of output channels.

\paragraph{Base network.} In this work we introduced EfficientDet~\cite{tan2020efficientdet} as base architectures, motivated not only by its performance in COCO-2017 but also by its reduced encoder size and compact tensor output at the end of each block. Compared to the Mask RCNN employed by Matsubara \etal~\cite{matsubara2020split} (7.8~GMAC, GMAC = $10^9$ multiply-accumulate operations), or the YOLO-v3 employed by Cohen \etal~\cite{cohen2020lightweight} (10.4~GMAC), EfficientDet (with scaling parameter $D=2$) has better mAP@50:95 on COCO-2017, with a much more economic encoder (2.4~GMAC). A silhouette comparison of the three architectures, showing only their major blocks, appears in Figure~\ref{fig:split_object_detectors}. We split the network at the latest possible point before branching, where the representation to be transmitted to the remote server is most compressible.

\paragraph{Choice of compressor.} The compressive autoencoder inserted at the split point is what allows a full-neural architecture.  Many alternatives appear in existing split networks, mainly with feed-forward convolutional layers (without skip connections), and ReLU activation~\cite{eshratifar2019bottlenet, eshratifar2019towards, matsubara2020split}, sigmoid activation~\cite{shao2020bottlenet++}, pointwise convolutions for channel reduction~\cite{eshratifar2019bottlenet, eshratifar2019towards}, and PReLU + Generalized Divisive Normalization (GDN)~\cite{jankowski2020joint}. 

The feedforward autoencoders implemented in Mask RCNN and Faster RCNN by Matsubara \etal~\cite{matsubara2020split} were unfeasible for our model, because, by themselves, they spent a great deal of computation (more than 6~GMAC), which is more than twice our \textit{entire encoder}. 

Experimental work (which we showcase in our experiments) shows a certain latitude in compressor choice. We tested a compressor inspired by Bottlenet~\cite{eshratifar2019bottlenet}, but without the non-differentiable JPEG layer, for we aimed at a full-neural solution: compressor and decompressor had a spatial reduction unit and a channel-wise reduction unit (both explained in their paper). In another scheme, we simply duplicated the last layer of EfficentDet twice (once as a compressor, once as a decompressor), changing only the \textbf{size of the bottleneck representation} $C$. We also tested a variation of the latter without the compressor layer, connecting the encoder’s output (again changing the size of the output bottleneck representation) directly on the decompressor. We tested different values for $C$, but contrarily to the freely configurable $\alpha$, changing $C$ requires retraining the model.


\subsection{Training}
\label{sec:training}

Our training procedure (Figure~\ref{fig:training_method}) merges the best advantages of Generalized Head Network Distillation~\cite{matsubara2021neural} and of MutualNet~\cite{yang2020mutualnet}. 

From the latter, we employ the slimmable training~\cite{yu2019universally} using several versions of the same set of weights, at $N$ widths at a time, choosing for each training step the lowest, the highest, and $N-2$ randomly chosen intermediate widths among all desired configuration widths for the network~\cite{yu2019universally}. Yu \etal call that procedure “sandwich rule”. The $N$ versions are used in a round-robin loop to compute the gradients from the loss and, ultimately, to update the single weight-set (see Algorithm~1 in~\cite{yu2019universally} ).


From the former, we adopt the loss based upon minimizing an $\ell_2$ norm between the \textit{features} extracted by student and teacher at several points of the decoder, including the point where the network splits. That procedure changes from MutualNet (and from  Yu \etal’s), which is based on a KL-divergence loss over the model outputs. 



As hinted above, one important parameter during training is the set of widths for which the network should learn to operate, e.g., ${\alpha} = {0.25, 0.33, 0.5, 0.66, 1.0}$. The number of widths may be larger than $N$, as different values will be chosen at random by the sandwich rule. In our experiments, we found out that $\alpha_{\mathrm{min}}$, the \textbf{minimum value in the set}, dominates the impact on performance.



\paragraph{Post-training batch normalization statistics.} A fine but crucial point in moving from Faster RCNN and Mask RCNN toward EfficientDet was \textit{removing} the post-training batchnorm statistics used by Yu \& Huang~\cite{yu2019universally}, since in our setting they considerably hindered the model.

\section{Experiments and results}
\label{sec:experiments}
\subsection{Materials and methods}
\label{sec:materials_and_methods}

We performed all experiments on the COCO-2017 dataset~\cite{lin2014microsoft}, measuring the outcomes on the validation split with the mAP@50:95 metric, the primary metric for the COCO challenge, which corresponds to the mean average precision, averaged for all region IoU thresholds between 50 and 95\%, with a 5\% step. We used the official COCO API to compute the metric.


We used the Faster~RCNN and Mask~RCNN models of Matsubara \etal~\cite{matsubara2021neural}, including the public trained weights, as non-configurable baselines. We contrast those models with the performance of four bandwidth-configurable models (Faster~RCNN, Mask~RCNN, EfficientDet-D1, and EfficientDet-D2) and two computation- and bandwidth-configurable models (EfficientDet-D2 and Faster~RCNN). 

We trained (distilled) all models for 12 epochs, using the largest batch size possible for each model (6 for non-configurable, 8 for configurable models). We hand-optimized the learning rates and schedulers of all models, setting learning rates to decrease by half every 3 epochs for non-configurable and bandwidth-configurable models, and every 2 epochs for full-configurable models. For inference, we quantized at 8 bits the output of the compressor for all models. Detailed procedures are available in our source repository\footnote{https://github.com/jsiloto/adaptive-cob}.


We implemented all models in PyTorch~\cite{paszke2017automatic}, from various open-source code-bases\footnote{https://github.com/zylo117/Yet-Another-EfficientDet-Pytorch}\footnote{https://github.com/JiahuiYu/slimmable\_networks}\footnote{https://github.com/yoshitomo-matsubara/hnd-ghnd-object-detectors}. For counting arithmetic operations, we used the ptflops Python library\footnote{ https://github.com/sovrasov/flops-counter.pytorch} (the ones for YOLO-v3 came from their code repository\footnote{https://github.com/pjreddie/darknet/issues/1039}).

\begin{figure}
\begin{center}
\includegraphics[width=0.99\linewidth]{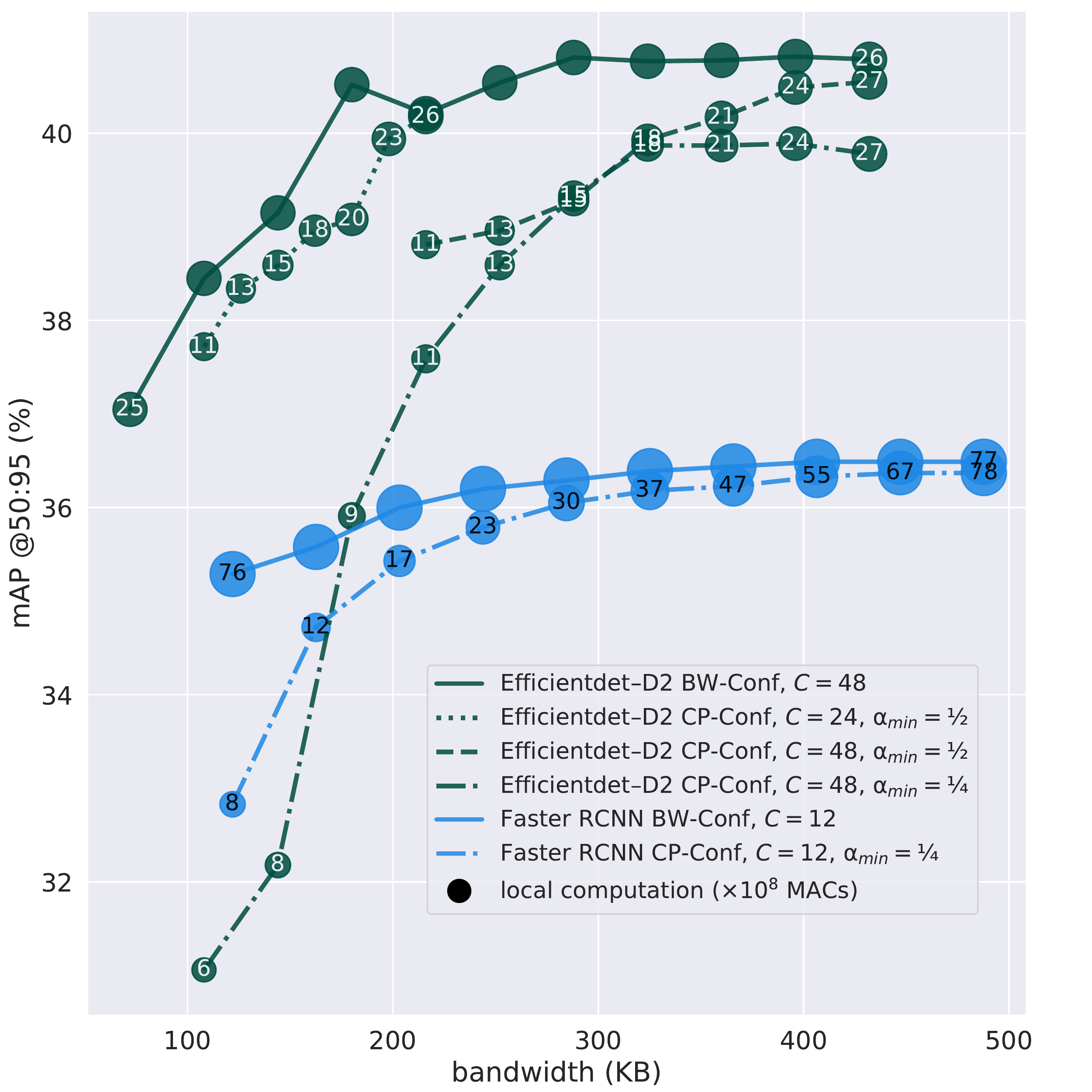}
\end{center}
   \caption{Comparison between models that are only bandwidth-configurable (BW-Conf) with models that are also computation-configurable (CP-Conf), showcasing local computation expenditure (multiply-accumulate operations on encoder) in the data points' sizes and labels. Remark that some configurations may present a similar compromise of bandwidth \textit{vs.} mAP but vastly different computation cost.} 
   \label{fig:graph_bw_map_macs}
\end{figure}

\subsubsection{Main results}
The comparison of the bandwidth-configurable techniques appears in Figure~\ref{fig:graph_bw_map}. Those results are easier to interpret keeping in mind the training procedure shown in Figure~\ref{fig:training_method}: the performance of the reference teacher networks appear as a single filled dot in the graph. The teacher's mAP, marked with the dotted line, gives the expected upper bound for the distilled students. The hollow dots show the baseline networks, learned by pure distillation: those networks are learned using the technique of Matsubara \etal~\cite{matsubara2021neural} and are \textit{not} configurable. 

Our results (half-filled dots) show that it is possible to make Mask RCNN and Faster RCNN configurable without impacting their performance. However, EfficientDet-D1 and, especially, EfficientDet-D2 present much better mAP $\vs.$ bandwidth compromises  — although there is some performance penalty in making them configurable. The penalty is constant and relatively small up to a certain bandwidth threshold, and then it grows steeply. Still, a single architecture (EfficientNet-D2) with a single set of weights is able to dominate a large area of the configuration space, essentially everything above the yellow dotted line. 

A finer-grained comparison of some of the models appears in Figure~\ref{fig:graph_bw_map_macs}, where we added full-configurable (computation and bandwidth) techniques. The small labels inside the dots indicate the computation spent in the encoder, in tenths of GMAC = $10^8$ multiply-accumulate operations. The areas of the dots also indicate, visually, that number. Here, the same bandwidth-only-configurable Faster RCNN and EfficientDet-D2 of Figure~\ref{fig:graph_bw_map} appear as baselines (continuous lines). To reduce clutter, we only labeled the first and last dots in those essentially computation-constant models. In the other series, we plot the results by varying the size of the bottleneck $C$ and the minimum value for $\alpha$ (see Figures~\ref{fig:proposed_architecture} and~\ref{fig:training_method}). Figure~\ref{fig:graph_bw_map_macs} shows that full-configuration is achievable, with considerable reduction in computation, and modest impact on mAP, except, again, below a certain threshold. Achieving full configuration, however, is less straightforward than bandwidth configuration. Contrarily to the bandwidth-only-configurable Efficient-D2, a single full-configurable version cannot be stretched over extended regimens of bandwidth and computation without suffering large penalties: notice how the left tail of the $(C=48, \alpha_{\mathrm{min}}=0.25)$ series drops sharply in mAP. Still, if covering such an extended range of configurations is necessary, it is possible to do it advantageously with just two sets of weights, instead of several. 

\subsubsection{Other analyses}

\begin{figure}
\begin{center}
\includegraphics[width=0.99\linewidth]{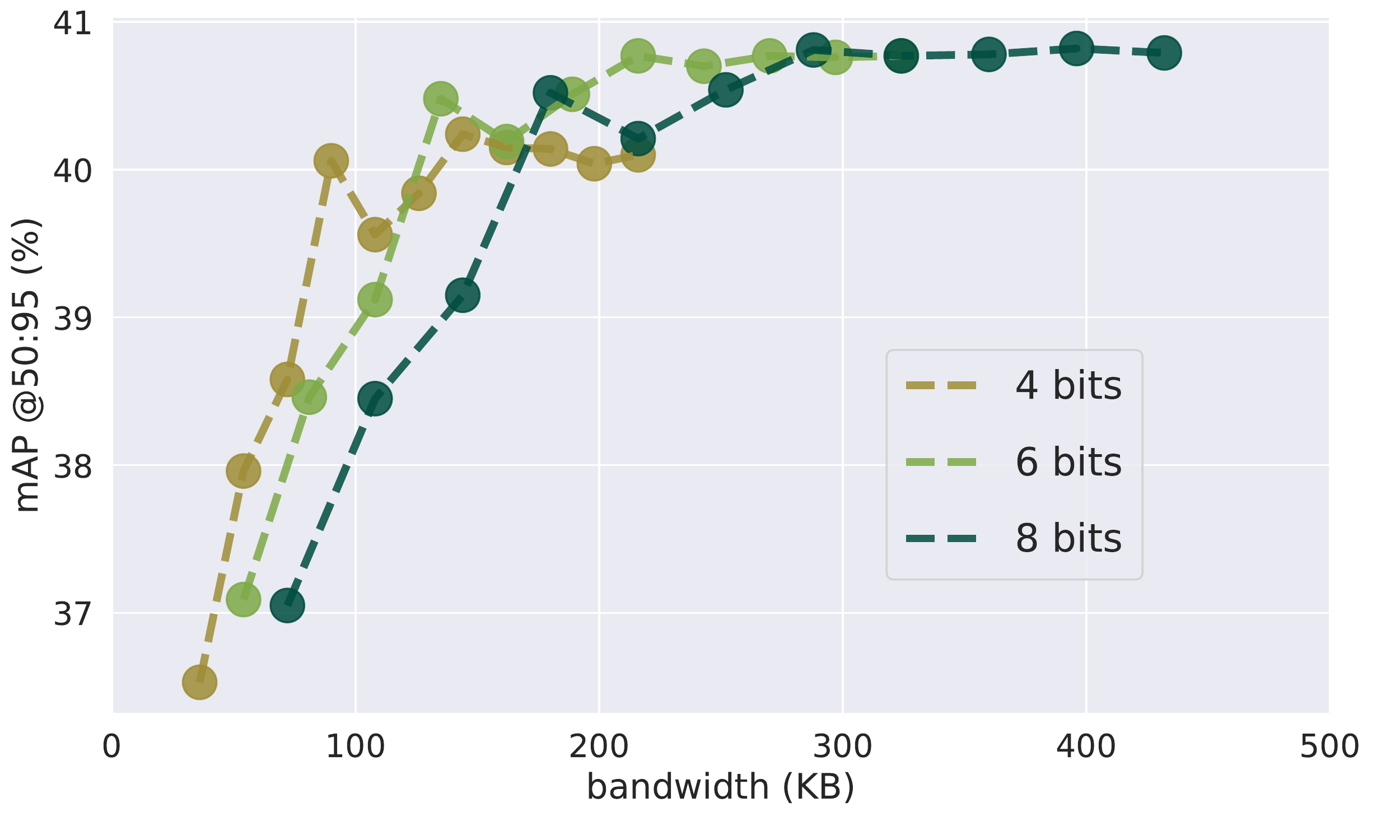}
\end{center}
   \caption{Interaction of the proposed technique with feature quantization: even aggressive quantization has a relative modest impact on mAP. Pure quantization does not provide the best operational compromise, but helps to improve the configurability of the proposed model. The model in dark green is the bandwidth-configurable EfficientDet-D2 that appears in Figure~\ref{fig:graph_bw_map}.}
\label{fig:graph_quantization}
\end{figure}

Since simple feature quantization is so cheap, it is interesting to analyze how it compares with our technique. As shown in Figure~\ref{fig:graph_quantization}, in terms of bandwidth, the effect of simple quantization is highly predictable, and may be computed deterministically (e.g., the bandwidth of a 4-bit-quantized compressor is exactly half that of 8-bit-quantized compressor). In terms of mAP, the results are not deterministic, but it is well known that deep networks are surprisingly robust to strong levels of quantization~\cite{carvalho2016deep, choi2018near}.  The results in the plot show that pure quantization does not always provide the best operational compromise, but that adding quantization to the mix may greatly improve the configurability of the model.

Many implementation choices proved to be more or less indifferent, demonstrating a certain robustness of the design. For example, several compressor choices resulted in very similar mAPs on EfficientDet (Table~\ref{table:compressor_ablation}). For the sake of uniformity, we opted for the design based on the last layer of EfficientDet for both the compressor and decompressor, but it is interesting to note that the design with \textit{only} the decompressor has very slightly lower performance. 
Other choices that had little impact were the use of a pretrained encoder \textit{vs.} random weights (Table~\ref{table:pretrained_ablation}) or enabling \textit{vs.} disabling stochastic depth~\cite{huang2016deep} (Table~\ref{table:sthocastic_depth_ablation}).

\begin{figure}
\begin{center}
\includegraphics[width=0.99\linewidth]{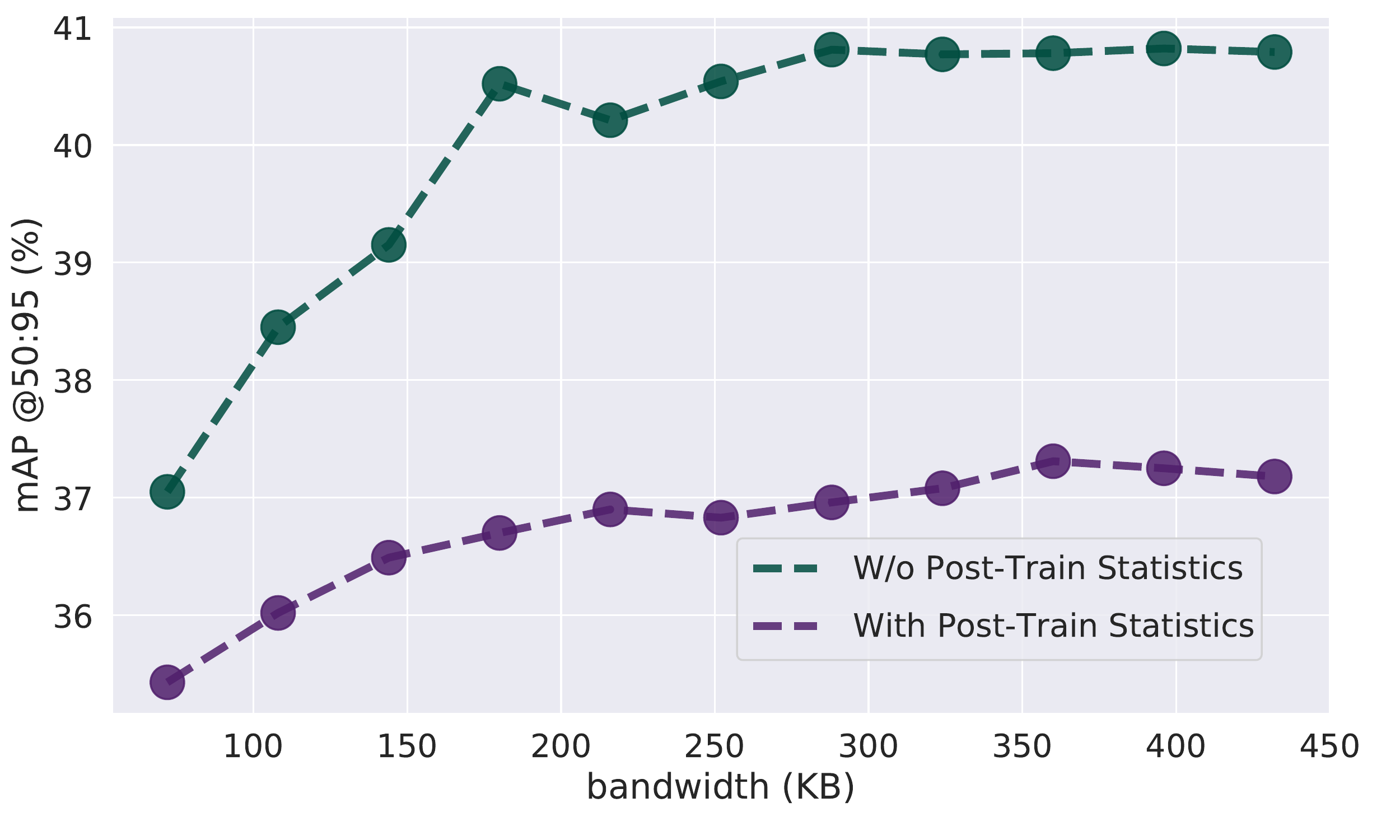}
\end{center}
   \caption{The post-train statistics for batch normalization, which were helpful for Mark RCNN and Faster RCNN, proved deleterious for EfficientDet. The model in dark green is the bandwidth-configurable EfficientDet-D2 that appears in Figure~\ref{fig:graph_bw_map}.}
\label{fig:graph_pbn_ablation}
\end{figure}

Other implementation “details” had considerable impact. The post-train statistics for batch normalization introduced by Yu \& Huang~\cite{yu2019universally}, with a positive effect on Mask RCNN and Faster RCNN, has a \textit{negative} effect on EfficientDet, as illustrated in Figure~\ref{fig:graph_pbn_ablation}.



\begin{table}
\begin{center}
\begin{tabular}{lc}
\toprule
Compressor $\longrightarrow$ Decompressor & mAP@50:95 \\
\midrule
SRU+CRU+ReLU $\longrightarrow$ SRU+CRU  & 40.57\% \\
EfficientDet LL $\longrightarrow$ EfficientDet LL & 40.53\%  \\
none $\longrightarrow$ EfficientDet LL & 40.44\% \\
\bottomrule
\end{tabular}
\end{center}
\caption{(De)compressors based on Spatial Reduction Unit, Channel Reduction Unit, EfficientDet Last Layer.}
\label{table:compressor_ablation}
\end{table}

\begin{table}
\begin{center}
\begin{tabular}{lc}
\toprule
Pretrained Encoder & mAP@50:95  \\
\midrule
Yes & 40.53\% \\
No & 40.23\% \\
\bottomrule
\end{tabular}
\end{center}
\caption{Impact of pretrained encoder.}
\label{table:pretrained_ablation}
\end{table}

\begin{table}
\begin{center}
\begin{tabular}{lc}
\toprule
Stochastic Depth & mAP@50:95 \\
\midrule
Yes  & 40.53\% \\
No & 40.53\% \\
\bottomrule
\end{tabular}
\end{center}
\caption{Impact of stochastic depth.}
\label{table:sthocastic_depth_ablation}
\end{table}

\section{Discussion}
\label{sec:discussion}
In order to reach large-scale deployment to consumers, deep learning models will have to adapt on-the-fly to many severe operational constraints. This work is a step in that direction, by allowing configurability of object detection for two operational constraints with a single weight-set. The experiments show that our design works for three very different base architectures (Mask RCNN, Faster RCNN, and EfficientDet), with robustness to small implementation details, like the choice of compressor.

For simplicity of the concept, in this work we pick a single slimming factor $\alpha$ for the entire encoder. In future works, we intend to explore whether two or more factors may be advantageous — in particular, decoupling the choices of computation and bandwidth. In our current model, the choice of quantization may help to decouple those dimensions, but several slimming factors may further enhance that decoupling.

Another simplifying decision was keeping a single set of weights for both encoder and decoder, but having more than one model on the remote decoder, which runs on a large server, may be feasible, and possibly advantageous. On the other direction, bandwidth and computation configurability may be relevant on the server in high-throughput situations to avoid queuing. Analyzing how client and server may interact on-the-fly to maintain optimal quality-of-service is a fascinating frontier for deep computer vision.

\section{Acknowledgments}
\label{sec:acknowledgments}
J. S. Assine is funded by the Coordenação de Aperfeiçoamento de Pessoal de Nível Superior (CAPES), Finance Code 001. Eduardo Valle is partially funded by CNPq Grant 315168/2020-0. J.C.S. Santos Filho is partially funded by CNPq Grant 306241/2019-6. The RECOD lab is funded by grants from FAPESP, CAPES, and CNPq.

{\small
\bibliographystyle{ieee_fullname}
\bibliography{egbib}

\begin{thebibliography}{10}\itemsep=-1pt

\bibitem{achille2018emergence}
Alessandro Achille and Stefano Soatto.
\newblock Emergence of invariance and disentanglement in deep representations.
\newblock {\em The Journal of Machine Learning Research}, 19(1):1947--1980,
  2018.

\bibitem{alvar2020bit}
Saeed~Ranjbar Alvar and Ivan~V Baji{\'c}.
\newblock Bit allocation for multi-task collaborative intelligence.
\newblock In {\em ICASSP 2020-2020 IEEE International Conference on Acoustics,
  Speech and Signal Processing (ICASSP)}, pages 4342--4346. IEEE, 2020.

\bibitem{assine2019compressing}
Juliano~S Assine, Alan Godoy, and Eduardo Valle.
\newblock Compressing representations for embedded deep learning.
\newblock {\em arXiv preprint arXiv:1911.10321}, 2019.

\bibitem{bajic2021collaborative}
Ivan~V Baji{\'c}, Weisi Lin, and Yonghong Tian.
\newblock Collaborative intelligence: Challenges and opportunities.
\newblock In {\em ICASSP 2021-2021 IEEE International Conference on Acoustics,
  Speech and Signal Processing (ICASSP)}, pages 8493--8497. IEEE, 2021.

\bibitem{balle2016end}
Johannes Ball{\'e}, Valero Laparra, and Eero~P Simoncelli.
\newblock End-to-end optimized image compression.
\newblock {\em arXiv preprint arXiv:1611.01704}, 2016.

\bibitem{cai2020once}
Han Cai, Chuang Gan, Tianzhe Wang, Zhekai Zhang, and Song Han.
\newblock Once-for-all: Train one network and specialize it for efficient
  deployment.
\newblock {\em arXiv preprint arXiv:1908.09791}, 2020.

\bibitem{callegaro2020optimal}
Davide Callegaro, Yoshitomo Matsubara, and Marco Levorato.
\newblock Optimal task allocation for time-varying edge computing systems with
  split dnns.
\newblock In {\em GLOBECOM 2020-2020 IEEE Global Communications Conference},
  pages 1--6. IEEE, 2020.

\bibitem{carvalho2016deep}
Micael Carvalho, Matthieu Cord, Sandra Avila, Nicolas Thome, and Eduardo Valle.
\newblock Deep neural networks under stress.
\newblock In {\em 2016 IEEE International Conference on Image Processing
  (ICIP)}, pages 4443--4447. IEEE, 2016.

\bibitem{chen2019lossy}
Zhuo Chen, Kui Fan, Shiqi Wang, Ling-Yu Duan, Weisi Lin, and Alex Kot.
\newblock Lossy intermediate deep learning feature compression and evaluation.
\newblock In {\em Proceedings of the 27th ACM International Conference on
  Multimedia}, pages 2414--2422, 2019.

\bibitem{choi2018deep}
Hyomin Choi and Ivan~V Baji{\'c}.
\newblock Deep feature compression for collaborative object detection.
\newblock In {\em 2018 25th IEEE International Conference on Image Processing
  (ICIP)}, pages 3743--3747. IEEE, 2018.

\bibitem{choi2018near}
Hyomin Choi and Ivan~V Baji{\'c}.
\newblock Near-lossless deep feature compression for collaborative
  intelligence.
\newblock In {\em 2018 IEEE 20th International Workshop on Multimedia Signal
  Processing (MMSP)}, pages 1--6. IEEE, 2018.

\bibitem{choi2020back}
Hyomin Choi, Robert~A Cohen, and Ivan~V Baji{\'c}.
\newblock Back-and-forth prediction for deep tensor compression.
\newblock In {\em ICASSP 2020-2020 IEEE International Conference on Acoustics,
  Speech and Signal Processing (ICASSP)}, pages 4467--4471. IEEE, 2020.

\bibitem{cohen2020lightweight}
Robert~A Cohen, Hyomin Choi, and Ivan~V Baji{\'c}.
\newblock Lightweight compression of neural network feature tensors for
  collaborative intelligence.
\newblock In {\em 2020 IEEE International Conference on Multimedia and Expo
  (ICME)}, pages 1--6. IEEE, 2020.

\bibitem{eshratifar2019jointdnn}
Amir~Erfan Eshratifar, Mohammad~Saeed Abrishami, and Massoud Pedram.
\newblock Jointdnn: an efficient training and inference engine for intelligent
  mobile cloud computing services.
\newblock {\em IEEE Transactions on Mobile Computing}, 2019.

\bibitem{eshratifar2019bottlenet}
Amir~Erfan Eshratifar, Amirhossein Esmaili, and Massoud Pedram.
\newblock Bottlenet: A deep learning architecture for intelligent mobile cloud
  computing services.
\newblock In {\em 2019 IEEE/ACM International Symposium on Low Power
  Electronics and Design (ISLPED)}, pages 1--6. IEEE, 2019.

\bibitem{eshratifar2019towards}
Amir~Erfan Eshratifar, Amirhossein Esmaili, and Massoud Pedram.
\newblock Towards collaborative intelligence friendly architectures for deep
  learning.
\newblock In {\em 20th International Symposium on Quality Electronic Design
  (ISQED)}, pages 14--19. IEEE, 2019.

\bibitem{horowitz20141}
Mark Horowitz.
\newblock 1.1 computing's energy problem (and what we can do about it).
\newblock In {\em 2014 IEEE International Solid-State Circuits Conference
  Digest of Technical Papers (ISSCC)}, pages 10--14. IEEE, 2014.

\bibitem{hu2020fast}
Diyi Hu and Bhaskar Krishnamachari.
\newblock Fast and accurate streaming cnn inference via communication
  compression on the edge.
\newblock In {\em 2020 IEEE/ACM Fifth International Conference on
  Internet-of-Things Design and Implementation (IoTDI)}, pages 157--163. IEEE,
  2020.

\bibitem{huang2016deep}
Gao Huang, Yu Sun, Zhuang Liu, Daniel Sedra, and Kilian~Q Weinberger.
\newblock Deep networks with stochastic depth.
\newblock In {\em European conference on computer vision}, pages 646--661.
  Springer, 2016.

\bibitem{huang2012close}
Junxian Huang, Feng Qian, Alexandre Gerber, Z~Morley Mao, Subhabrata Sen, and
  Oliver Spatscheck.
\newblock A close examination of performance and power characteristics of 4g
  lte networks.
\newblock In {\em Proceedings of the 10th international conference on Mobile
  systems, applications, and services}, pages 225--238, 2012.

\bibitem{embedded_market_by_2019}
Press~Release IMARC.
\newblock Embedded systems market size is growing at 5.6\% cagr to reach 95400
  million usd in 2024, Feb 2019.

\bibitem{jankowski2020joint}
Mikolaj Jankowski, Deniz G{\"u}nd{\"u}z, and Krystian Mikolajczyk.
\newblock Joint device-edge inference over wireless links with pruning.
\newblock In {\em 2020 IEEE 21st International Workshop on Signal Processing
  Advances in Wireless Communications (SPAWC)}, pages 1--5. IEEE, 2020.

\bibitem{lin2014microsoft}
Tsung-Yi Lin, Michael Maire, Serge Belongie, James Hays, Pietro Perona, Deva
  Ramanan, Piotr Doll{\'a}r, and C~Lawrence Zitnick.
\newblock Microsoft coco: Common objects in context.
\newblock In {\em European conference on computer vision}, pages 740--755.
  Springer, 2014.

\bibitem{liu2020improve}
Jianhui Liu and Qi Zhang.
\newblock To improve service reliability for ai-powered time-critical services
  using imperfect transmission in mec: An experimental study.
\newblock {\em IEEE Internet of Things Journal}, 7(10):9357--9371, 2020.

\bibitem{luo2020comparison}
Chunjie Luo, Xiwen He, Jianfeng Zhan, Lei Wang, Wanling Gao, and Jiahui Dai.
\newblock Comparison and benchmarking of ai models and frameworks on mobile
  devices.
\newblock {\em arXiv preprint arXiv:2005.05085}, 2020.

\bibitem{luo2018deepsic}
Sihui Luo, Yezhou Yang, Yanling Yin, Chengchao Shen, Ya Zhao, and Mingli Song.
\newblock Deepsic: Deep semantic image compression.
\newblock In {\em International Conference on Neural Information Processing},
  pages 96--106. Springer, 2018.

\bibitem{matsubara2019distilled}
Yoshitomo Matsubara, Sabur Baidya, Davide Callegaro, Marco Levorato, and Sameer
  Singh.
\newblock Distilled split deep neural networks for edge-assisted real-time
  systems.
\newblock In {\em Proceedings of the 2019 Workshop on Hot Topics in Video
  Analytics and Intelligent Edges}, pages 21--26, 2019.

\bibitem{matsubara2020head}
Yoshitomo Matsubara, Davide Callegaro, Sabur Baidya, Marco Levorato, and Sameer
  Singh.
\newblock Head network distillation: Splitting distilled deep neural networks
  for resource-constrained edge computing systems.
\newblock {\em IEEE Access}, 8:212177--212193, 2020.

\bibitem{matsubara2020split}
Yoshitomo Matsubara and Marco Levorato.
\newblock Split computing for complex object detectors: Challenges and
  preliminary results.
\newblock In {\em Proceedings of the 4th International Workshop on Embedded and
  Mobile Deep Learning}, pages 7--12, 2020.

\bibitem{matsubara2021neural}
Yoshitomo Matsubara and Marco Levorato.
\newblock Neural compression and filtering for edge-assisted real-time object
  detection in challenged networks.
\newblock In {\em 2020 25th International Conference on Pattern Recognition
  (ICPR)}, pages 2272--2279. IEEE, 2021.

\bibitem{paszke2017automatic}
Adam Paszke, Sam Gross, Soumith Chintala, Gregory Chanan, Edward Yang, Zachary
  DeVito, Zeming Lin, Alban Desmaison, Luca Antiga, and Adam Lerer.
\newblock Automatic differentiation in pytorch.
\newblock 2017.

\bibitem{shao2020bottlenet++}
Jiawei Shao and Jun Zhang.
\newblock Bottlenet++: An end-to-end approach for feature compression in
  device-edge co-inference systems.
\newblock In {\em 2020 IEEE International Conference on Communications
  Workshops (ICC Workshops)}, pages 1--6. IEEE, 2020.

\bibitem{shao2020communication}
Jiawei Shao and Jun Zhang.
\newblock Communication-computation trade-off in resource-constrained edge
  inference.
\newblock {\em IEEE Communications Magazine}, 58(12):20--26, 2020.

\bibitem{shi2020communication}
Yuanming Shi, Kai Yang, Tao Jiang, Jun Zhang, and Khaled~B Letaief.
\newblock Communication-efficient edge ai: Algorithms and systems.
\newblock {\em IEEE Communications Surveys \& Tutorials}, 22(4):2167--2191,
  2020.

\bibitem{tan2020efficientdet}
Mingxing Tan, Ruoming Pang, and Quoc~V Le.
\newblock Efficientdet: Scalable and efficient object detection.
\newblock In {\em Proceedings of the IEEE/CVF Conference on Computer Vision and
  Pattern Recognition}, pages 10781--10790, 2020.

\bibitem{teerapittayanon2016branchynet}
Surat Teerapittayanon, Bradley McDanel, and Hsiang-Tsung Kung.
\newblock Branchynet: Fast inference via early exiting from deep neural
  networks.
\newblock In {\em 2016 23rd International Conference on Pattern Recognition
  (ICPR)}, pages 2464--2469. IEEE, 2016.

\bibitem{tishby2015deep}
Naftali Tishby and Noga Zaslavsky.
\newblock Deep learning and the information bottleneck principle.
\newblock In {\em 2015 IEEE Information Theory Workshop (ITW)}, pages 1--5.
  IEEE, 2015.

\bibitem{wang2020convergence}
Xiaofei Wang, Yiwen Han, Victor~CM Leung, Dusit Niyato, Xueqiang Yan, and Xu
  Chen.
\newblock Convergence of edge computing and deep learning: A comprehensive
  survey.
\newblock {\em IEEE Communications Surveys \& Tutorials}, 22(2):869--904, 2020.

\bibitem{yang2020resolution}
Le Yang, Yizeng Han, Xi Chen, Shiji Song, Jifeng Dai, and Gao Huang.
\newblock Resolution adaptive networks for efficient inference.
\newblock In {\em Proceedings of the IEEE/CVF Conference on Computer Vision and
  Pattern Recognition}, pages 2369--2378, 2020.

\bibitem{yang2020mutualnet}
Taojiannan Yang, Sijie Zhu, Chen Chen, Shen Yan, Mi Zhang, and Andrew Willis.
\newblock Mutualnet: Adaptive convnet via mutual learning from network width
  and resolution.
\newblock In {\em European Conference on Computer Vision}, pages 299--315.
  Springer, 2020.

\bibitem{yang2017designing}
Tien-Ju Yang, Yu-Hsin Chen, and Vivienne Sze.
\newblock Designing energy-efficient convolutional neural networks using
  energy-aware pruning.
\newblock In {\em Proceedings of the IEEE Conference on Computer Vision and
  Pattern Recognition}, pages 5687--5695, 2017.

\bibitem{yu2019universally}
Jiahui Yu and Thomas~S Huang.
\newblock Universally slimmable networks and improved training techniques.
\newblock In {\em Proceedings of the IEEE/CVF International Conference on
  Computer Vision}, pages 1803--1811, 2019.

\bibitem{Yu2019}
Jiahui Yu, Linjie Yang, Ning Xu, Jianchao Yang, and Thomas Huang.
\newblock {Slimmable neural networks}.
\newblock In {\em 7th International Conference on Learning Representations,
  ICLR 2019}, 2019.

\end{thebibliography}
}

\end{document}